\documentclass[10pt, a4paper]{article}
\PassOptionsToPackage{table}{xcolor}
\usepackage[final]{lrec2026} 

\usepackage{times}
\usepackage{latexsym}
\usepackage{makecell}
\usepackage{threeparttablex}
\usepackage{amssymb}   
\usepackage{xcolor}
\definecolor{lightgrayrow}{gray}{0.93}   
\definecolor{lightgraycell}{gray}{0.93}
\usepackage{stfloats}
\usepackage{placeins}

\usepackage[T1]{fontenc}

\usepackage[utf8]{inputenc}

\usepackage{microtype}

\usepackage{inconsolata}
\usepackage{float}  
\usepackage{graphicx}
\usepackage{comment}
\usepackage{booktabs}
\usepackage{multirow}
\usepackage{ragged2e}
\usepackage{listings}
\usepackage{xcolor}
\usepackage{tabularx}  
\usepackage{subcaption}
\usepackage{amsmath}

\lstdefinelanguage{json}{
  basicstyle=\ttfamily\footnotesize,
  stringstyle=\color{red},
  keywordstyle=\color{blue},
  commentstyle=\color{gray},
  morestring=[b]",
  morekeywords={true, false, null},
}

\newcommand{\SummaryGPT}{\textit{GPTSum}}
\newcommand{\SummaryLLaMA}{\textit{LLaMASum}}
\newcommand{\GSnip}{\textit{GSnip}}
\newcommand{\GPTFourMini}{\textsc{GPT-4o} mini}
\newcommand{\LLaMAThree}{\textsc{LLaMA 3.1--8B Instruct}}

\title{Dynamically Acquiring Text Content to Enable the Classification of Lesser-known Entities for Real-world Tasks}

\name{Fahmida Alam, Ellen Riloff} 

\address{University of Arizona, Tucson, AZ, USA \\
         \{fahmidaalam, riloff\}@arizona.edu\\}

\abstract{
Existing Natural Language Processing (NLP) resources often lack the task-specific information required for real-world problems and provide limited coverage of lesser-known or newly introduced entities. For example, business organizations and health care providers may need to be classified into a variety of different taxonomic schemes for specific application tasks. Our goal is to enable domain experts to easily create a task-specific classifier for entities by providing only entity names and gold labels as training data. Our framework then dynamically acquires descriptive text about each entity, which is subsequently used as the basis for producing a text-based classifier. We propose a novel text acquisition method that leverages both web and large language models (LLMs). We evaluate our proposed framework on two classification problems in distinct domains: (i) classifying organizations into Standard Industrial Classification (SIC) Codes\textsuperscript{1}, which categorize organizations based on their business activities; and (ii) classifying healthcare providers into healthcare provider taxonomy codes\textsuperscript{2}, which represent a provider’s medical specialty and area of practice. Our best-performing model achieved macro-averaged F1-scores of 82.3\% and 72.9\% on the SIC code and healthcare taxonomy code classification tasks, respectively.
\\
\newline 
}

\begin{document}

\maketitleabstract
\setcounter{footnote}{1}
\footnotetext{The U.S. government established Standard Industrial Classification (SIC) codes to classify organizations based on the activities they serve and operate within.}
\setcounter{footnote}{2}
\footnotetext{The Health Care Provider Taxonomy code set, maintained by the National Uniform Claim Committee (NUCC), classifies health care providers by grouping, classification, and area of specialization, such as Cardiology, Obstetrics \& Gynecology, or Family Medicine.}

\section{Introduction}
Many real-world applications require knowledge about named entities, including organizations, people, and places. To address this need, researchers have developed structured data resources, such as knowledge bases and knowledge graphs, that compile information about a wide range of named entities. DBpedia \cite{10.5555/1785162.1785216}, Freebase \cite{10.1145/1376616.1376746}, Wikidata \cite{10.1145/2629489}, and Yago \cite{10.1145/1242572.1242667} are examples of widely used structured knowledge resources. These resources typically capture attributes of general interest, such as the location and size of organizations or the professions of individuals, making them broadly useful.

However, many real-world applications require entity-specific knowledge that is not present in existing resources. For instance, applications may need to classify organizations by their business activities, revenue sources, client types, or ownership structures. Furthermore, categorization schemes can vary significantly across applications. For example, business activities could be characterized by a small number of high-level categories or broken down into a large number of fine-grained categories that are relevant to the application domain. 

Our goal is to develop a method that can rapidly acquire new types of information about named entities given only their names as input, without the need for an explicit text corpus or structured knowledge resource to enable the learning. We introduce a framework that acquires descriptive text for entities given only their names as input and subsequently trains classifiers using the acquired text along with the corresponding gold labels. For text acquisition, the framework integrates web retrieval and LLM-based generation to automatically produce descriptive text for each entity. The framework is adaptable, scalable, and requires minimal human supervision, supporting the rapid development of entity classification systems across diverse tasks and domains. To demonstrate its generalizability, we applied the framework to two classification tasks from entirely different domains, and the results highlight its effectiveness and robustness. All in all, our contributions are:


\begin{enumerate}
    \item We propose a generalizable framework that takes only the entity names and their corresponding gold labels as input. It handles the entire process through a novel text acquisition method that leverages web retrieval and LLM-based generation to produce descriptive text for classifier training. This approach eliminates dependence on pre-compiled datasets containing entity descriptions, which represents the key novelty of our work.
    
    \item We evaluate our framework on two different types of real-world classification tasks: (i) classifying organizations into Standard Industrial Classification (SIC) codes and (ii) classifying healthcare providers into healthcare provider taxonomy codes.

    \item We constructed two benchmark datasets using our framework in two distinct domains: (i) industry and (ii) healthcare, to demonstrate its effectiveness and generalizability. Evaluation results indicate that our framework achieves robust performance across domains. 
    We release both datasets on GitHub\footnote{\url{https://github.com/alamfahmida/dynamic-text-acquisition-entity-classification}}  to facilitate future research in automated knowledge acquisition, text-based classification, and entity classification, while following responsible data-sharing practices.

\end{enumerate}

\section{Related Work}
Named entity recognition and entity classification have been extensively studied in NLP, but have traditionally focused on labeling entity mentions in a document or text fragment (e.g., \cite{article,10.5555/2900728.2900742,10.1145/2187836.2187898,yaghoobzadeh-schutze-2015-corpus,10184827}). In contrast, our research aims to acquire knowledge about real-world entities irrespective of any specific mention or document. 

Our research shares the same high-level goal as Knowledge Base Population (KBP), which extracts information to populate structured knowledge bases, such as Wikidata, DBpedia, or TAC-KBP \cite{10.5555/2002472.2002618,Surdeanu2013OverviewOT}. Prior work has utilized weak or distant supervision by leveraging structured knowledge resources like Wikipedia or Freebase to acquire training data (e.g., \cite{mintz-etal-2009-distant,Riedel2010ModelingRA,hoffmann-etal-2011-knowledge}). However, our approach does not use any data from structured knowledge resources or external corpora. Our goal is to acquire knowledge that is not currently available in existing resources, to augment them or populate new ones.

Although our method utilizes retrieved text, it differs fundamentally from Retrieval-Augmented Generation (RAG) approaches, which typically retrieve documents from a static, indexed corpus, such as Wikipedia, The Pile \cite{gao2020pile800gbdatasetdiverse}, or similar sources, to guide a generative model at inference time \cite{10.5555/3495724.3496517,10.5555/3524938.3525306,izacard-grave-2021-leveraging}. In contrast, our approach performs web search to proactively retrieve context and leverages multiple LLMs to generate task-specific text. These sources, individually and in combination, are then used to construct training data for multiple classifiers trained under a supervised learning paradigm.
Some prior work has integrated search engine retrieval into language models \cite{karpukhin-etal-2020-dense, DBLP:journals/corr/abs-2107-07566}, but such approaches are primarily designed for document ranking or generation tasks, rather than training a classifier to assign labels based on retrieved content. To the best of our knowledge, no prior work has combined web-retrieved and LLM-generated text as complementary sources to construct training data for supervised classification.

Our work was inspired by the research of \cite{jiang2023classifyingorganizationsfoodontologies}, which used google-retrieved text to classify organizations for knowledge-graph population in the food systems domain. However, our framework goes beyond web retrieval by also leveraging multiple LLMs for task-specific text generation, exploring whether LLMs can reduce dependence on existing structured knowledge sources for various NLP tasks. Moreover, while their experiments were limited to BERT-based models, we evaluate our framework using three additional language models, RoBERTa, Longformer, and \GPTFourMini{}, to assess performance across different model families. Furthermore, we experimented with confidence-based thresholds to produce high-precision predictions, enabling the automatic annotation of new training data with minimal human intervention.

\section{Task Definition and Dataset}
\label{sec:task-def}
Although our experiments focus on the following tasks, the proposed framework is task-agnostic and can be adapted to a wide range of entity-centric categorization and knowledge acquisition tasks.

\subsection{SIC Code}
\paragraph{Task Definition} In the SIC code classification task, organizations are categorized by their Standard Industrial Classification (SIC) codes, which describe their primary business activities and are assigned by the U.S. government.\footnote{These codes are analogous to North American Industry Classification System (NAICS) codes used across North America.} SIC codes are widely used for economic analyses and food supply chain studies \cite{jiang2023classifyingorganizationsfoodontologies}. These codes follow a hierarchical structure: a four-digit code specifies an industry nested within broader categories. For example, the code 0116 represents the Soybeans industry under Cash Grains, within Agricultural Production Crops. This hierarchy enables analysis at varying levels of granularity by grouping organizations by the first one to four digits.

\paragraph{Dataset} As a starting point, we adopt the dataset introduced by \citelanguageresource{jiang2023sicdataset}, originally developed for food system research, and enrich it with task-relevant texts harvested using our framework. The organizations in this dataset primarily consist of lesser-known organizations rather than widely recognized corporations like Google or IBM. Most of them, such as Multi-Corp International Inc., Fonon Corp., and JMXI Inc., are not commonly mentioned in general resources.\footnote{The dataset released on GitHub includes the complete list of organizations.} Following their setup, we use the first two digits of the SIC codes as category labels to reduce sparsity and capture each organization’s broad business activities. The dataset contains 5,400 organizations labeled with two-digit SIC codes across 27 categories (see Table~\ref{tab:task-categories}). The dataset is partitioned into 2,700 training, 900 development, and 1,800 test instances.
\begin{table}[ht]
  \centering
  \small
  \renewcommand{\arraystretch}{1.15}
  \setlength{\tabcolsep}{6pt}
  \begin{tabular}{p{0.90\columnwidth}}
    \Xhline{0.4pt}
    SIC Code Categories \\
    \Xhline{0.3pt}
    Metal Mining; Oil and Gas Extraction; Food and Kindred Products; Printing, Publishing and Allied Industries; Chemicals and Allied Products; Fabricated Metal Products; Industrial and Commercial Machinery; Electronic; Transportation Equipment; Measuring, Photographic, Medical, Communications; Electric, Gas and Sanitary Services; Wholesale Trade - Durable Goods; Wholesale Trade - Nondurable Goods; Eating and Drinking Places; Miscellaneous Retail; Depository Institutions; Nondepository Credit Institutions; Security; Insurance Carriers; Real Estate; Holding and Other Investment Offices; Hotels, Rooming Houses, Camps; Business Services; Amusement and Recreation Services; Health Services; Engineering, Accounting, Research \\[4pt]
    \Xhline{0.3pt}
    Healthcare Taxonomy Code Categories \\
    \Xhline{0.3pt}
    Allopathic \& Osteopathic Physicians; Behavioral Health and Social Service Providers; Chiropractic Providers; Dental Providers; Dietary and Nutritional Service Providers; Emergency Medical Service Providers; Eye and Vision Service Providers; Nursing Service Providers; Nursing Service Related Providers; Other Service Providers; Pharmacy Service Providers; Physician Assistants and Advanced Practice Nursing Providers; Podiatric Medicine and Surgery Service Providers; Respiratory, Developmental, Rehabilitative and Restorative Service Providers; Speech, Language and Hearing Service Providers; Student, Health Care; Technologists, Technicians, and Other Technical Service Providers \\
    \Xhline{0.3pt}
  \end{tabular}
  \caption{Categories used in the SIC code (27 categories) and healthcare provider taxonomy code (17 categories) classification tasks.}
  \label{tab:task-categories}
  \vspace{-1em}
\end{table}


\subsection{Healthcare Provider Taxonomy Code}
\paragraph{Task Definition}The healthcare provider taxonomy classification task involves assigning healthcare professionals to their corresponding taxonomy codes, which define their medical specialty and area of practice. Each taxonomy code is a ten-character alphanumeric identifier from the Health Care Provider Taxonomy Code Set maintained by the National Uniform Claim Committee (NUCC)\footnote{\url{https://nucc.org/}}. The code follows a three-level hierarchy: (i) Provider Grouping, a broad category such as Allopathic and Osteopathic Physicians; (ii) Classification, a discipline within the group, e.g., Internal Medicine; and (iii) Specialization, a focused area like Cardiovascular Disease. For example, \texttt{207RC0000X} represents the grouping \texttt{Allopathic and Osteopathic Physicians}, classification \texttt{Internal Medicine}, and specialization \texttt{Cardiovascular Disease}. Thus, by identifying a provider’s taxonomy code, one can infer these specific details about their professional domain.

\paragraph{Dataset} To start, we took provider names and their corresponding gold categories from the National Plan and Provider Enumeration System \citelanguageresource{nppes}\footnote{The National Plan and Provider Enumeration System (NPPES) is a public registry maintained by the Centers for Medicare \& Medicaid Services (CMS). It includes information on U.S. healthcare providers, their specialty taxonomy codes, and National Provider Identifiers (NPIs). Links: \href{https://www.nber.org/research/data/national-plan-and-provider-enumeration-system-nppesnpi}{NPPES}, \url{https://www.cms.gov/}.}
. As mentioned earlier, the 10-digit taxonomy follows a three-level hierarchy, but the NUCC documentation does not specify digit boundaries for each level. Therefore, we use the full 10-digit taxonomy codes. The code set covers individuals, groups, and non-individual entities. Since we focus on healthcare providers’ taxonomy code, we target only the \texttt{individual category}, which contains information exclusively about individual healthcare providers. There are a total of 17 fixed categories (see Table~\ref{tab:task-categories}), each containing multiple taxonomy codes. For instance, the category \textit{Allopathic \& Osteopathic Physicians} includes codes such as \texttt{207K00000X} and \texttt{2081S0010X}, etc.

To construct a balanced subset across the taxonomy hierarchy, we focused on 17 individual categories and selected one taxonomy code from each category that had at least 200 individual provider instances in the NPPES database. Then we randomly sampled 200 individual providers, resulting in a total of 3,400 instances. We then split the dataset into training, development, and test sets using the same ratio as in our SIC classification task, yielding 1,700 training, 1,140 test, and 560 development instances. 



\section{Proposed Framework}

\begin{figure}[H]
  \centering
  \includegraphics[width=\columnwidth]{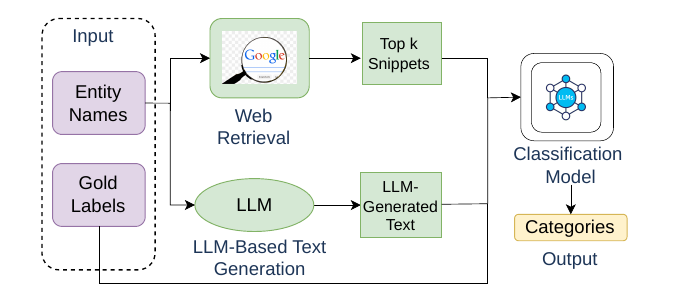}
\caption{Overview of the proposed architecture. The input consists of entity names and their corresponding gold labels. The framework employs two components for text acquisition: (i) a web retrieval module that retrieves top-\emph{k} snippets, and (ii) an LLM-based module that generates task-specific descriptive text. The retrieved and generated texts, along with their gold labels, are then used to train a classification model that predicts the task-specific category for a given entity.}
  \label{fig:KH_architecture}
\end{figure}

We propose a framework that takes an entity as input and autonomously handles the entire process, from text acquisition to classifier training. For training purposes, the framework requires gold labels for the training set, which are not used during text acquisition. Figure~\ref{fig:KH_architecture} illustrates the components and workflow of our architecture. The framework consists of two main steps: (1) \textsc{Text Acquisition} and (2) \textsc{ Classification Model Training}. 

\subsection{Text Acquisition}
\label{sec:text-harvesting}

The key idea behind our approach is that entities can be categorized using text descriptions that are {\it automatically} acquired for the specific application task, without relying on any pre-compiled text resources. We combine two approaches to obtain task-relevant text for each entity: (1) retrieving text using a search engine and (2) generating text with large language models. For the SIC code classification task, ``task-relevant'' text refers to descriptions of an organization’s business activities, whereas for the healthcare taxonomy code classification task, it refers to information about a provider’s medical specialty, professional focus, and area of practice.

\subsubsection{Google Snippets}
\begin{figure}[H]
\vspace{-1em}
\centering
\includegraphics[width=\columnwidth]{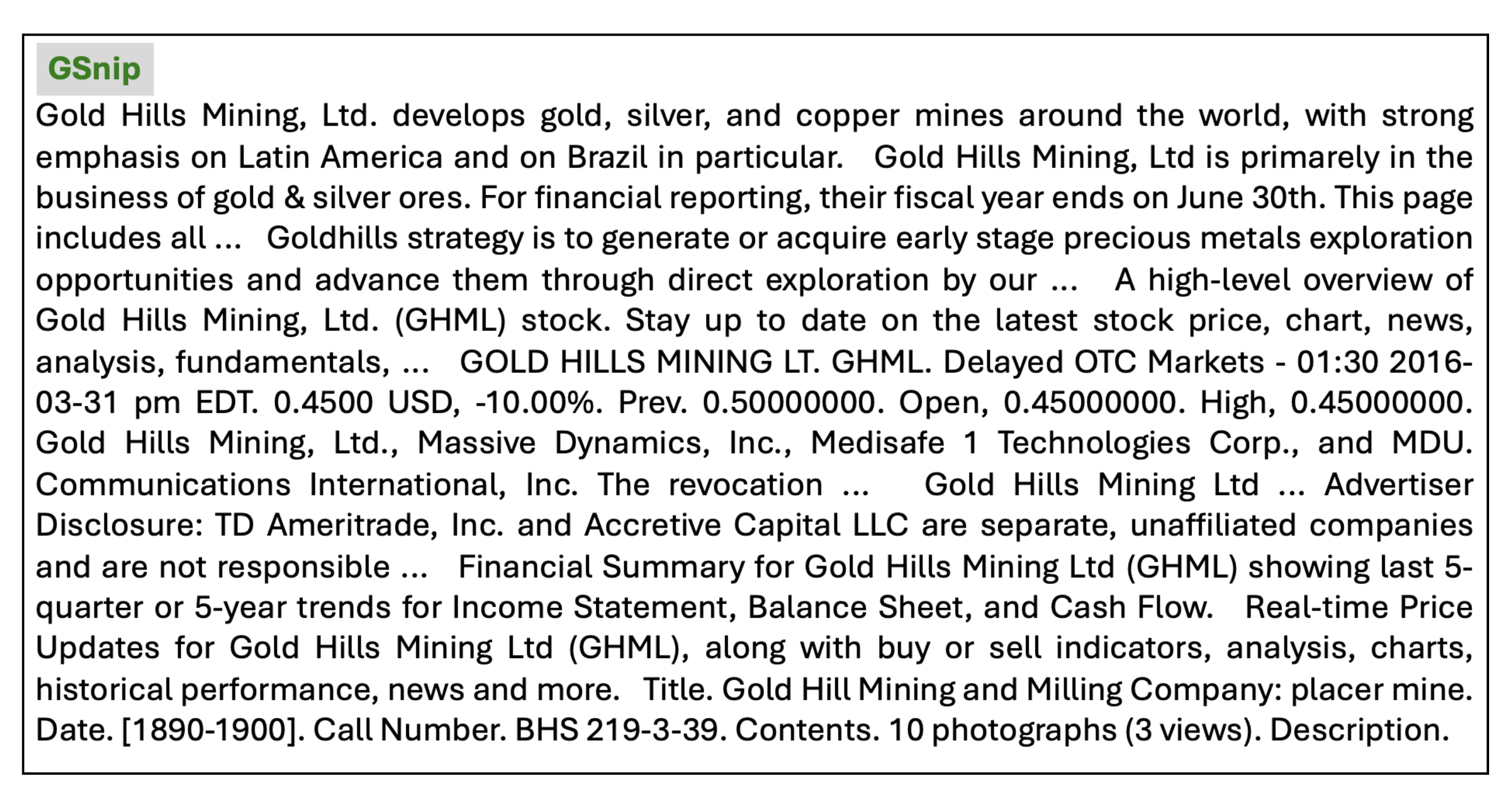}
    \vspace{-1.5em}
    \caption{Example \GSnip{} for organization \textit{Gold Hills Mining, Ltd.}.}
  \label{fig:GSnip}
  \vspace{-1em}
\end{figure}

We used the SerpAPI\footnote{\url{https://serpapi.com/}} interface to programmatically query the Google Search Engine\footnote{\url{https://programmablesearchengine.google.com/about/}}
 for each entity name (organization or healthcare provider), without quotes. For each query, we considered the top 10 search results. In our experiments, we used text snippets, which are small blocks of text that appear underneath a link to a webpage in a search engine results page. These snippets are typically around 100–200 characters long and provide a short description of the webpage content. We obtained the snippets from the top 10 retrieved results and concatenated them into a single text block. We refer to this aggregated text as \textit{GSnip}, which serves as semantically rich content capturing how the entity is described across multiple sources. Figure~\ref{fig:GSnip} shows the {\it GSnip} text acquired for the organization \textit{Gold Hills Mining, Ltd.}.

\subsubsection{LLM-Generated Summaries}
\label{sec: LLM-Generated Summaries}
We also generated task-specific summaries using two large language models (LLMs): \GPTFourMini{}\footnote{Specifically, we used {\tt gpt-4o-mini-2024-07-18}.} \citelanguageresource{gpt} and \LLaMAThree{}\footnote{\url{https://huggingface.co/meta-llama/Llama-3.1-8B-Instruct}} \citelanguageresource{llama}. These summaries aim to concisely capture the key characteristics of each entity. By comparing summaries generated by different language models, we assess the consistency and semantic quality of LLM-generated text descriptions. Due to architectural and alignment differences between models, we adopted distinct prompting strategies. 

\begin{figure}[h]  
  \centering
\includegraphics[width=\columnwidth]{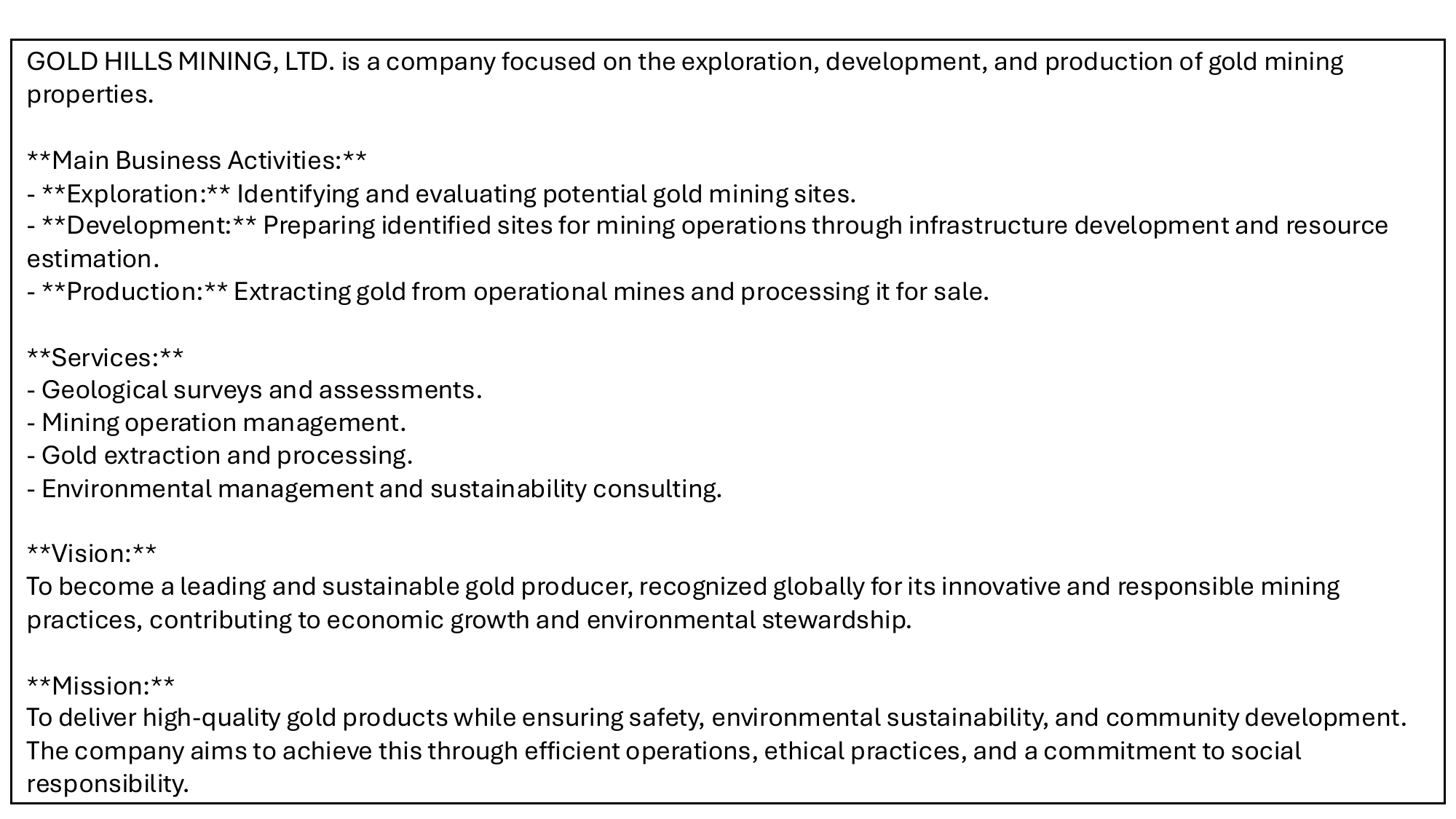}
    \vspace{-2em} 
    \caption{Example of \SummaryGPT{} for \textit{Gold Hills Mining, Ltd.}.}
  \label{fig:GPTSum_exampl}
\end{figure}

\paragraph{GPT-generated Summaries}
For the SIC code task, we prompted the model with: \texttt{``Summarize the main business activities, services, vision, and mission of [ORG\_NAME].''}. 
For the healthcare taxonomy code task, the prompt was: \texttt{``Summarize the healthcare specialization, scope of practice, and typical services provided by [Provider\_NAME]. The summary should describe the clinician’s professional type and main field of practice, following standard U.S. healthcare taxonomy conventions.''}. We refer to the resulting summary as \SummaryGPT{}. The aggregated Google Snippets (GSnip) are typically around 250–300 words, so we set the model's maximum token limit to produce summaries of comparable length. Figure~\ref{fig:GPTSum_exampl} shows an example of \SummaryGPT{} for \textit{Gold Hills Mining, Ltd.}.

\begin{figure}[h]  
  \centering
\includegraphics[width=\columnwidth]{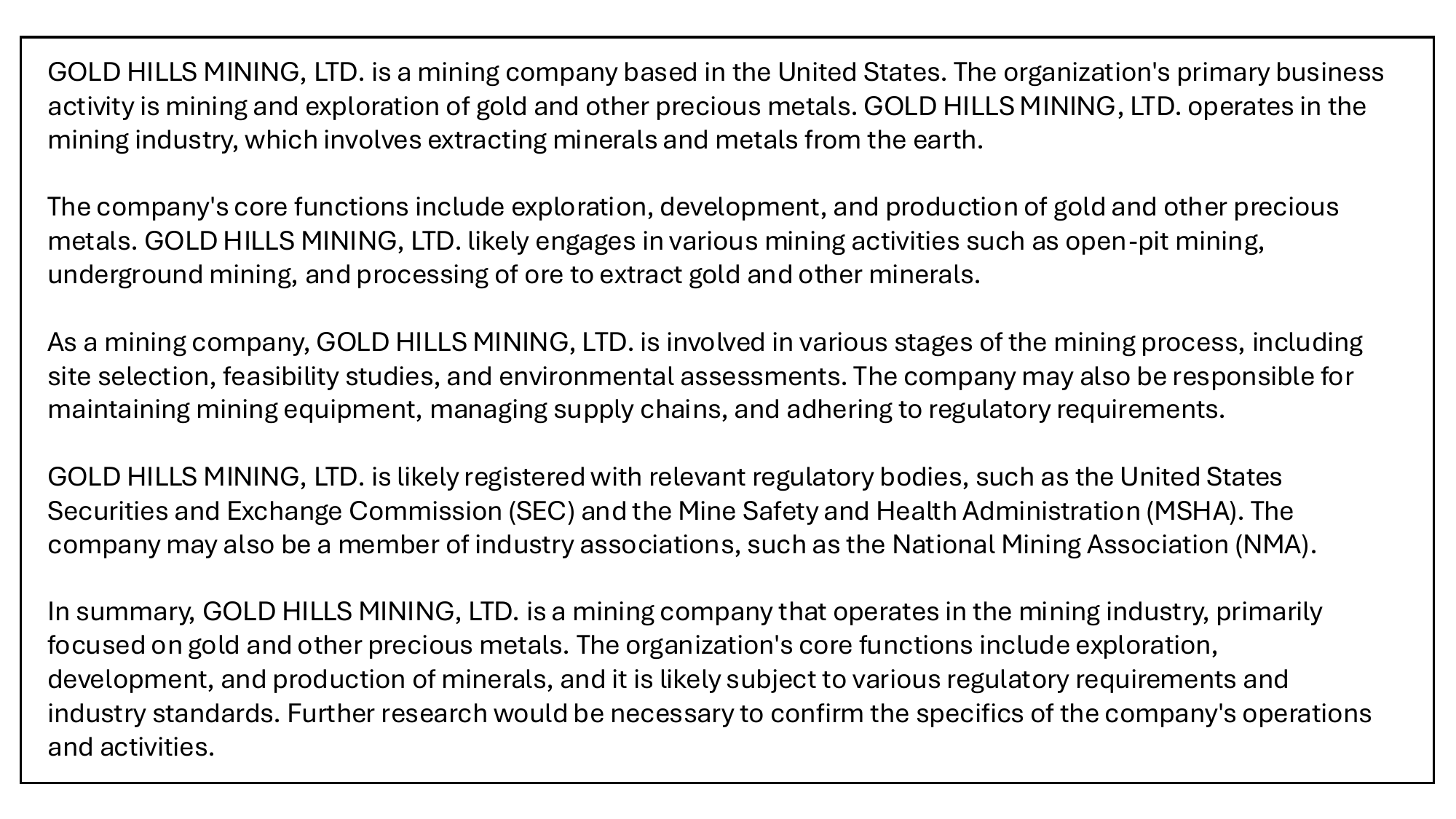}
    \vspace{-2em} 
    \caption{Example of \SummaryLLaMA{} for \textit{Gold Hills Mining, Ltd.}.}
  \label{fig:llamaSum_exampl}
\end{figure}

\paragraph{LLaMA-generated Summaries}
Being smaller and more sensitive to prompt specificity, \textsc{LLaMA 3.1--8B Instruct} benefited from more explicit instruction designed to constrain text generations to factual and verifiable content. This helped reduce hallucination and ensured consistency in structure and coverage with GPT-generated summaries. For the SIC code task, the prompt was: \texttt{``You are an assistant writing a factual summary about an organization based on its name. Given the [ORG\_NAME], your goal is to identify and describe the organization's main business activities, core functions, and the industry it operates in. Use only publicly verifiable information. The description should be informative, objective, and around 250–300 words. Do not add any assumptions or speculative content.''}. 
For the healthcare taxonomy task, the prompt was: \texttt{``You are a research assistant writing a factual summary about a healthcare provider’s specialty. Given the [PROVIDER\_NAME], your goal is to identify and describe their medical specialty, professional focus, qualifications, and the healthcare sector they operate in. Use only publicly verifiable information. The description should be informative, objective, and around 250--300 words. Do not add any assumptions or speculative content.''}. The resulting summary is referred to as \SummaryLLaMA. An example of \SummaryLLaMA{} is illustrated in the Figure~\ref{fig:llamaSum_exampl} for the entity \textit{Gold Hills Mining, Ltd.}.

\subsection{Classification Model Training}
\label{sec:classification-models}
We trained classification models by fine-tuning both encoder-based and generative language models. Each model was evaluated using five types of texts: (1) \GSnip{}, (2) \SummaryGPT{}, (3) \SummaryLLaMA{}, (4) \GSnip{} + \SummaryGPT{}, and (5) \GSnip{} + \SummaryLLaMA{}.
\subsubsection{Encoder-based Language Models}
We fine-tuned three encoder-based language models: (1) BERT \cite{DBLP:journals/corr/abs-1810-04805}, (2) RoBERTa \cite{DBLP:journals/corr/abs-1907-11692}, and (3) Longformer \cite{DBLP:journals/corr/abs-2004-05150}. Each model was trained on our training set using identical hyperparameters, which are described below. For each entity, the input consists of the entity name concatenated with its corresponding text content, and its gold category label. Hyperparameter values were selected on the basis of performance in the development set. 

\paragraph{BERT}
We fine-tuned the BERT model (\texttt{bert-base-uncased}) to create a classifier for our tasks. Input texts were tokenized using BERT's tokenizer with a maximum sequence length of 512 tokens. The final hidden state of the [CLS] token was fed into a linear classification layer to produce the predicted code.

\paragraph{RoBERTa}
We also fine-tuned the RoBERTa model (\texttt{roberta-base}), which shares the same transformer architecture as BERT but is pre-trained on a larger corpus using dynamic masking and without the Next Sentence Prediction objective. Input texts were tokenized using the RoBERTa tokenizer with a maximum sequence length of 512 tokens. As with BERT, classification was performed using the final hidden state of the first token (the [CLS] equivalent in RoBERTa).

\paragraph{Longformer}
We fine-tuned Longformer (\texttt{allenai/longformer-base-4096}) to assess whether its sparse attention mechanism offers any advantage in classification tasks. Inputs were tokenized using the Longformer tokenizer with a maximum sequence length of 1,024 tokens. The final hidden state of the [CLS] token configured with global attention was used as input to a linear classification layer. 

\paragraph{Hyperparameters for BERT, RoBERTa, and Longformer}
We fine-tuned all three models using the same hyperparameters: 3 training epochs, the AdamW optimizer, a learning rate of 5e-5, 500 warmup steps, and a weight decay of 0.01. We used a training batch size of 8 and an evaluation batch size of 16, and evaluated performance at the end of each epoch. 

\subsubsection{Generative Language Model}
\paragraph{\GPTFourMini{}}
We also fine-tuned \GPTFourMini{} (\texttt{gpt-4o-mini-2024-07-18}), a lightweight and cost-effective generative language model. We prepared training, dev, and test sets in an OpenAI-compatible chat format. The training and dev sets follow the same structure: each instance includes a system instruction and a user prompt consisting of an entity name concatenated with its acquired text description, paired with its gold label. In the test set, the gold label is omitted. Examples of the training and dev set formats are shown in Figure~\ref{fig:finetuneGPT-trainDev}, and the test format is shown in Figure~\ref{fig:finetuneGPT-inference} in Appendix~\ref{sec:appendix}. We followed the same format for the healthcare taxonomy classification task. The model was fine-tuned using supervised learning with default hyperparameters (\texttt{batch\_size}, \texttt{n\_epochs}, and \texttt{learning\_rate\_multiplier} set to \texttt{"auto"}).

\section{Experiments and Results}
In this section, we report the results on both tasks, evaluated using macro-averaged precision (P), recall (R), and F1-score.
\subsection{Prompting Baselines}
As a baseline, we experimented with prompting to determine whether state-of-the-art LLMs can effectively assign SIC categories to organizations and taxonomy codes to healthcare providers, which is important to assess whether the model has encountered such codes in its pre-training data. We used the \GPTFourMini{}\footnote{We used \texttt{gpt-4o-mini-2024-07-18} for these experiments.} language model for these experiments because of its strong performance-efficiency trade-off and suitability for large-scale prompting experiments at low cost.
We experimented with two prompting settings, one without additional context and another incorporating task-relevant text as context. For the no-context setting, the prompt was: \texttt{``You are a classification assistant for [TASK\_NAME]. Given the [ENTITY\_NAME] below, predict the [CODE\_TYPE] that best represents its [CATEGORY\_DESCRIPTION]. Choose ONLY one code from the provided options: [CODE\_LIST]. Return ONLY the code. Do not include explanations or extra text.''} 

For the with-context setting, we used the same prompt, except that additional context was included along with the entity name. As context, we used three types of text: \GSnip{}, \SummaryGPT{}, and \SummaryLLaMA{}. This experiment evaluates whether providing additional text as context can improve the LLM’s prompting performance. We used the same prompt structure for both the SIC code and healthcare provider taxonomy classification tasks, substituting the \texttt{[TASK\_NAME]}, \texttt{[ENTITY\_NAME]}, \texttt{[DESCRIPTION]}, \texttt{CATEGORY\_DESCRIPTION}, and \texttt{[CODE\_LIST]} accordingly. 
\subsubsection{Baseline Results}
\label{sec:prompting_noContext}
 
\begin{table}[t]
\centering
\small
\renewcommand{\arraystretch}{0.95}
\setlength{\tabcolsep}{2.6pt} 
\resizebox{\columnwidth}{!}{  
\begin{tabular}{@{\extracolsep{2pt}}lrrrrrr}
\toprule
\raisebox{-2ex}\GPTFourMini{} & \multicolumn{3}{c}{SIC Code} & \multicolumn{3}{c}{Healthcare Taxonomy} \\
\cmidrule(lr){2-4} \cmidrule(lr){5-7}
& P & R & F1 & P & R & F1 \\
\midrule
Prompting & 0.572 & 0.453 & 0.464 &  0.039 & 0.052 & 0.021 \\
\midrule
Prompting w/ \\[-1pt]
\hspace{3mm} \SummaryLLaMA{} & 0.536 & 0.466 & 0.464 & 0.023  & 0.058  & 0.010 \\
\hspace{3mm} \SummaryGPT{} & 0.565 & 0.496 & 0.497 & 0.111 & 0.055 & 0.046 \\
\hspace{3mm} \GSnip{} & \textbf{0.653} & {0.611} & \textbf{0.601} & \textbf{0.289} & \textbf{0.272} & \textbf{0.256} \\
\bottomrule
\end{tabular}
}
\vspace{-0.3em}
\caption{Prompting performance of \GPTFourMini{} across two classification tasks.}
\label{tab:gpt4o_prompting}
\vspace{-0.4em}
\end{table}

Table~\ref{tab:gpt4o_prompting} shows that prompting \GPTFourMini{} yields only a 46.4\% F1 score for organization-level SIC code classification, and just 2.1\% for healthcare provider taxonomy classification. These results suggest that \GPTFourMini{} has not memorized SIC or healthcare taxonomy codes from its pre-training data, indicating that this is not a trivial lookup task even for a large language model. 

Adding context to the prompt yields mixed outcomes, as shown in Table~\ref{tab:gpt4o_prompting}. For SIC code classification, \SummaryGPT{} improves F1 to 49.7\%, while \SummaryLLaMA{} offers no gain. Similar trends appear in the healthcare taxonomy task, where context slightly helps but overall accuracy remains low. Using \GSnip{} provides the largest boost, reaching 60.1\% and 25.6\% F1 for SIC code and healthcare taxonomy code classification, respectively. These results show that prompting alone is insufficient; however, adding context improves performance by complementing the model’s pre-trained knowledge. Nevertheless, the overall performance remains poor, indicating the need for new and more effective methods
\begin{table}[t]
\centering
\begin{ThreePartTable}
\small
\renewcommand{\arraystretch}{0.95}
\setlength{\tabcolsep}{2.8pt}
\resizebox{\columnwidth}{!}{
\begin{tabular}{@{\extracolsep{2pt}}lrrrrrr}
\toprule
\raisebox{-2ex}{Model\textsubscript{Context}} & \multicolumn{3}{c}{SIC Code} & \multicolumn{3}{c}{Healthcare Taxonomy} \\
\cmidrule(lr){2-4} \cmidrule(lr){5-7}
& P & R & F1 & P & R & F1 \\
\midrule
BERT \\[-1pt] 
\hspace{2mm} \SummaryLLaMA{} & 0.490 & 0.513 & 0.480 & 0.106 & 0.185 & 0.121 \\
\hspace{2mm} \SummaryGPT{} & 0.562 & 0.560 & 0.532 & 0.143 & 0.173 & 0.118 \\
\hspace{2mm} \GSnip{} \raisebox{0.0ex}{$^\bigstar$} & 0.700 & 0.699 & 0.699 & 0.555 & 0.508 & 0.485 \\
\hspace{2mm} \GSnip{} + \SummaryLLaMA{} & 0.723 & 0.728 & 0.699 & 0.723 & 0.662 & 0.672 \\
\hspace{2mm} \GSnip{} + \SummaryGPT{} & 0.724 & 0.718 & 0.700 & 0.738 & 0.679 & 0.687 \\
\cmidrule(lr){1-7}
RoBERTa \\[-1pt]
\hspace{2mm} \SummaryLLaMA{} & 0.525 & 0.512 & 0.483 & 0.096 & 0.175 & 0.092 \\
\hspace{2mm} \SummaryGPT{} & 0.581 & 0.565 & 0.550 & 0.222 & 0.189 & 0.139 \\
\hspace{2mm} \GSnip{} & 0.757 & 0.748 & 0.741 & 0.688 & 0.608 & 0.604 \\
\hspace{2mm} \GSnip{} + \SummaryLLaMA{} & 0.772 & 0.755 & 0.750 & 0.718 & 0.690 & 0.697 \\
\hspace{2mm} \GSnip{} + \SummaryGPT{} & 0.774 & 0.769 & 0.763 & 0.742 & 0.683 & 0.693 \\
\cmidrule(lr){1-7}
Longformer \\[-1pt]
\hspace{2mm} \SummaryLLaMA{} & 0.512 & 0.515 & 0.481 & 0.055 & 0.069 & 0.049 \\
\hspace{2mm} \SummaryGPT{} & 0.586 & 0.586 & 0.581 & 0.162 & 0.169 & 0.089 \\
\hspace{2mm} \GSnip{} & 0.754 & 0.754 & 0.746 & 0.707 & 0.634 & 0.643 \\
\hspace{2mm} \GSnip{} + \SummaryLLaMA{} & 0.740 & 0.727 & 0.716 & 0.714 & 0.685 & 0.691 \\
\hspace{2mm} \GSnip{} + \SummaryGPT{} & 0.767 & 0.758 & 0.750 & 0.693 & 0.689 & 0.694 \\
\cmidrule(lr){1-7}
GPT-4o-mini \\[-1pt]
\hspace{2mm} \SummaryLLaMA{} & 0.657 & 0.648 & 0.649 & 0.187 & 0.219 & 0.191 \\
\hspace{2mm} \SummaryGPT{} & 0.673 & 0.663 & 0.664 & 0.249 & 0.237 & 0.221 \\
\hspace{2mm} \GSnip{} & 0.823 & 0.818 & 0.817 & 0.742 & 0.725 & 0.728 \\
\hspace{2mm} \GSnip{} + \SummaryLLaMA{} & \textbf{0.827} & \textbf{0.825} & \textbf{0.823} & \textbf{0.745} & \textbf{0.726} & \textbf{0.729} \\
\hspace{2mm} \GSnip{} + \SummaryGPT{} & 0.826 & 0.822 & 0.822 & 0.717 & 0.703 & 0.705 \\
\bottomrule
\end{tabular}
}
\begin{minipage}{0.92\linewidth}
\vspace{1mm}
\tiny
$^\bigstar$~ Results for BERT trained with \GSnip{} for the SIC code classification task are taken from \cite{jiang2023classifyingorganizationsfoodontologies}. \\
\end{minipage}
\end{ThreePartTable}
\caption{Performance of fine-tuned models using different types of text on two classification tasks from distinct domains. (i) The \textit{SIC Code} section reports results for organization-level SIC code classification, and (ii) the \textit{Healthcare Taxonomy} section reports results for healthcare provider taxonomy code classification. Performance is measured using macro-averaged precision, recall, and F1-score. The best-performing model, \GPTFourMini{}, trained with \GSnip{}+\SummaryLLaMA{}, is shown in bold.} 
\label{tab:ft_with_gsnip_summary}
\end{table}

\subsection{Results of Our Framework}
Table \ref{tab:ft_with_gsnip_summary} presents the performance of four language models fine-tuned using different types of text. Across all architectures, BERT, RoBERTa, Longformer, and \GPTFourMini{}, the models trained with either (\GSnip{} + \SummaryGPT{}) or (\GSnip{} + \SummaryLLaMA{}) outperform those trained with individual text sources (\GSnip{}, \SummaryGPT{}, or \SummaryLLaMA{}). Our overall best-performing model is \GPTFourMini{}, trained with \GSnip{}+\SummaryLLaMA{}, bolded in Table~\ref{tab:ft_with_gsnip_summary}, achieving 82.3\% and 72.9\% macro-averaged F1 scores for the SIC code classification and healthcare taxonomy classification tasks, respectively. Even the simple BERT model trained with \GSnip{}+\SummaryGPT{} and \GSnip{}+\SummaryLLaMA{} outperforms BERT trained only with \GSnip{} by 20.2 and 18.7 points in the healthcare taxonomy classification task. For the SIC code classification task, the results are comparable. This trend confirms that combining retrieved and generated text provides complementary information that enhances classification performance regardless of the model type. Among the encoder-based models, RoBERTa with \GSnip{}+\SummaryGPT{} achieves a 76.3\% F1 score for the SIC code classification task, while Longformer achieves a 69.4\% F1 score for the healthcare taxonomy classification task. These represent the best-performing encoder-based models following \GPTFourMini{}.

When comparing Table~\ref{tab:gpt4o_prompting} and Table~\ref{tab:ft_with_gsnip_summary}, it is evident that fine-tuning with text acquired by our framework substantially outperforms all prompting-based approaches, demonstrating the overall efficiency of our framework. Our best-performing model (bolded in Table~\ref{tab:ft_with_gsnip_summary}) further outperforms the best prompting result, \GPTFourMini{} with \GSnip{}, by 22.2 points for the SIC code classification task and 47.3 points for the healthcare taxonomy classification task.

With \GSnip{}+\SummaryLLaMA{}, \GPTFourMini{} achieves the best F1 scores across both tasks, while \GSnip{}+\SummaryGPT{} shows a slight drop of 2.3 points in the healthcare taxonomy task compared to \GSnip{}. This suggests that when fine-tuned on its self-generated summaries, the model may become biased toward its prior knowledge, but by leveraging the strengths of both web-retrieved and LLM-generated text, the overall performance remains strong. Furthermore, the results remain robust irrespective of which LLM is used for summary generation. Both \SummaryGPT{} and \SummaryLLaMA{} contribute valuable information when combined with \GSnip{}, yielding improved performance across different models in both tasks.

These findings show that our framework is model-agnostic and summary-source-independent: it reliably improves classification performance whether we use GPT or LLaMA-based summaries and whether the underlying classifier is an encoder or a generative model. Overall, the consistent improvements across architectures and domains validate the effectiveness and generalizability of our proposed framework.




\section{Analysis}
We analyze the SIC code classification task as a representative case study to gain insights into the performance and design choices of our framework.

\subsection{Why do Google Snippets outperform LLM summaries?}

Our first analysis investigates why Google snippets performed better than  LLM-generated summaries (specifically, \SummaryGPT{}). We manually looked at 25 instances for which  \SummaryGPT{} led to an incorrect label but  \GSnip{} produced the correct label. 

In four cases, \GPTFourMini{} did not produce any summary (e.g., {\it ``I don’t have current detailed information \dots''}), leaving the model with no usable context so predictions were essentially arbitrary. 
However, Google returned information that supported the correct label. 
This finding shows that LLMs cannot always provide information about specific real-world entities if those entities are rare.

The most interesting issue pertains to the semantic framing of LLM-generated summaries. The summaries sometimes mentioned many things and included highly relevant information alongside tangential information. 
For example, \SummaryGPT{} emphasized secondary activities for \textit{SPARTON CORP}, such as medical instrumentation, resulting in a \textit{Measuring, Photographic, Medical (9)} label instead of the correct \textit{Electronic (7)} label. Similarly, \textit{AmREIT Monthly Income \& Growth Fund IV LP} was misclassified as \textit{Real Estate (20)} due to the absence of financial and investment context, which was present in \GSnip{} and led to the correct label \textit{Holding and Investment Offices (21)}.



Overall, the summaries produced by \GPTFourMini{} were coherent and fluent but often presented a broad narrative that omitted the most relevant operational information. In contrast, \textit{GSnip} returned Web content that usually focused on the most salient facts. Aggregating 10 snippets also provided more semantically diverse content that enhanced robustness. 


\subsection{Performance across Categories}
\begin{figure*}[t]
  \centering
  \includegraphics[width=\textwidth]{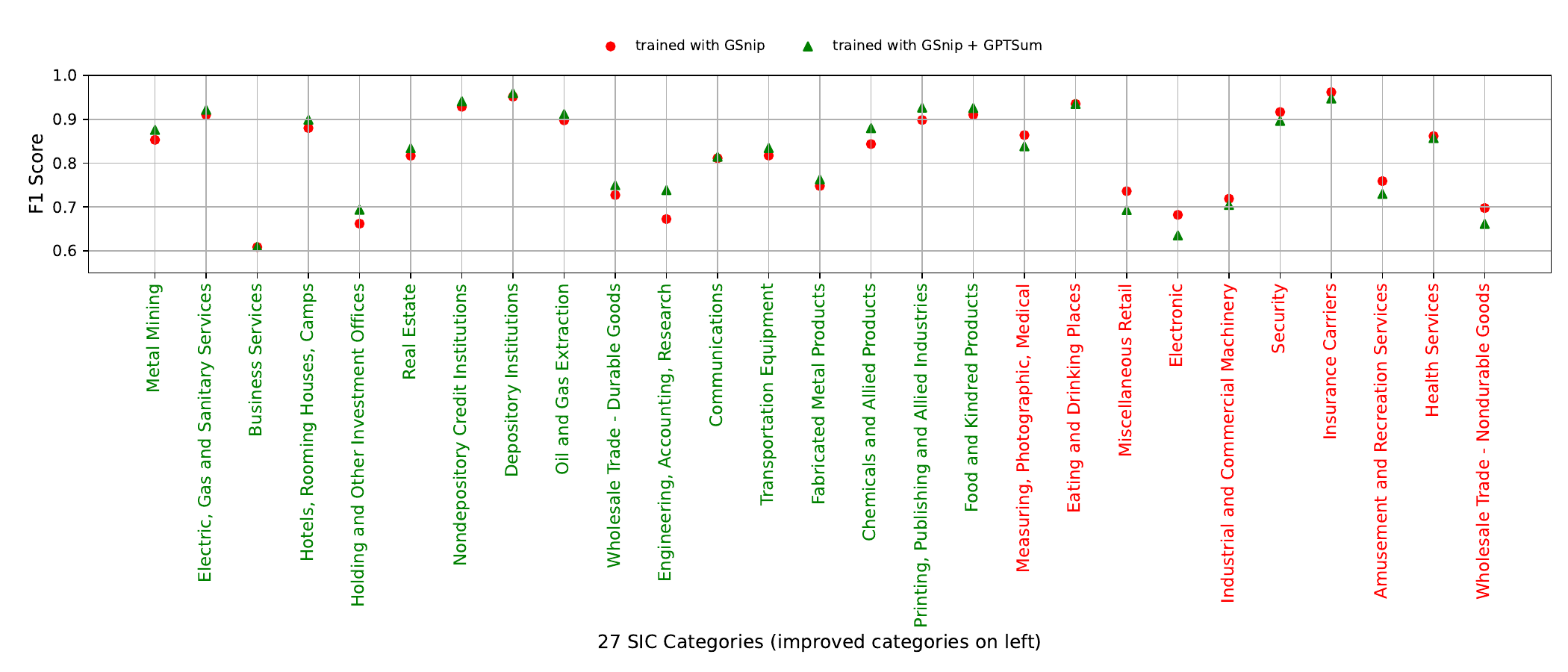}
  \caption{F1 scores across 27 SIC categories for \GPTFourMini{} fine-tuned with \GSnip{} vs. \GSnip{}+\SummaryGPT{}.}
  \label{fig:f1_comparison}
\end{figure*}

Figure~\ref{fig:f1_comparison} shows the F1 scores for each SIC category individually, for two of our models:  \GPTFourMini{} using \GSnip{} or using \GSnip{}+\SummaryGPT{}. The X-axis displays the categories, and the Y-axis shows the F1 scores. Each category is color-coded: green indicates categories where the combined text (\GSnip{}+\SummaryGPT{}) outperforms \GSnip{} alone, and red indicates where \GSnip{} yields better performance. These results show that 17 of the 27 categories benefitted from using both the LLM-generated summaries and the Google snippets. 

Overall, performance varies across categories. Some lower-performing categories are challenging due to their generality (e.g., \textit{Miscellaneous Retail} and \textit{Business Services}), while others require fine-grained semantic distinctions (e.g., there are distinct categories for Wholesale Trade of Durable vs. Nondurable Goods).

\subsection{Why Top-10 Google Snippets?}
We also evaluated performance using different numbers of Google snippets to assess their impact on the results.
Table~\ref{tab:gpt4o_ablationStudy} shows that using multiple snippets improves performance up to a point: the F1 score increases steadily from top-1 to top-10, peaking at \textbf{0.817}. Beyond that, additional snippets provide diminishing returns and may even degrade performance, likely due to noise or redundancy.
\begin{table}[h]
    \centering
    \small
    \renewcommand{\arraystretch}{1.2}
    \setlength{\tabcolsep}{8pt} 
    \begin{tabular}{lccc}
        \hline
        Fine-tuned \GPTFourMini{} & P & R & F1 \\
        \hline
        Top-1 snippet  & 0.691 & 0.687 & 0.686 \\
        Top-5 snippets  & 0.771 & 0.768 & 0.768 \\
        Top-10 snippets & \textbf{0.823} & \textbf{0.818} & \textbf{0.817} \\
        Top-15 snippets & 0.779 & 0.776 & 0.777 \\
        Top-20 snippets & 0.813 & 0.811 & 0.810 \\
        \hline
    \end{tabular}
    \caption{Performance of \GPTFourMini{} across different numbers of retrieved Google snippets.}
    \label{tab:gpt4o_ablationStudy}
    \vspace{-1em}
\end{table}

\subsection{High Precision Categorization}
\begin{figure}[t!]
  \vspace{-0.8em} 
  \centering
  \includegraphics[width=\columnwidth, trim=70 58 70 20, clip]{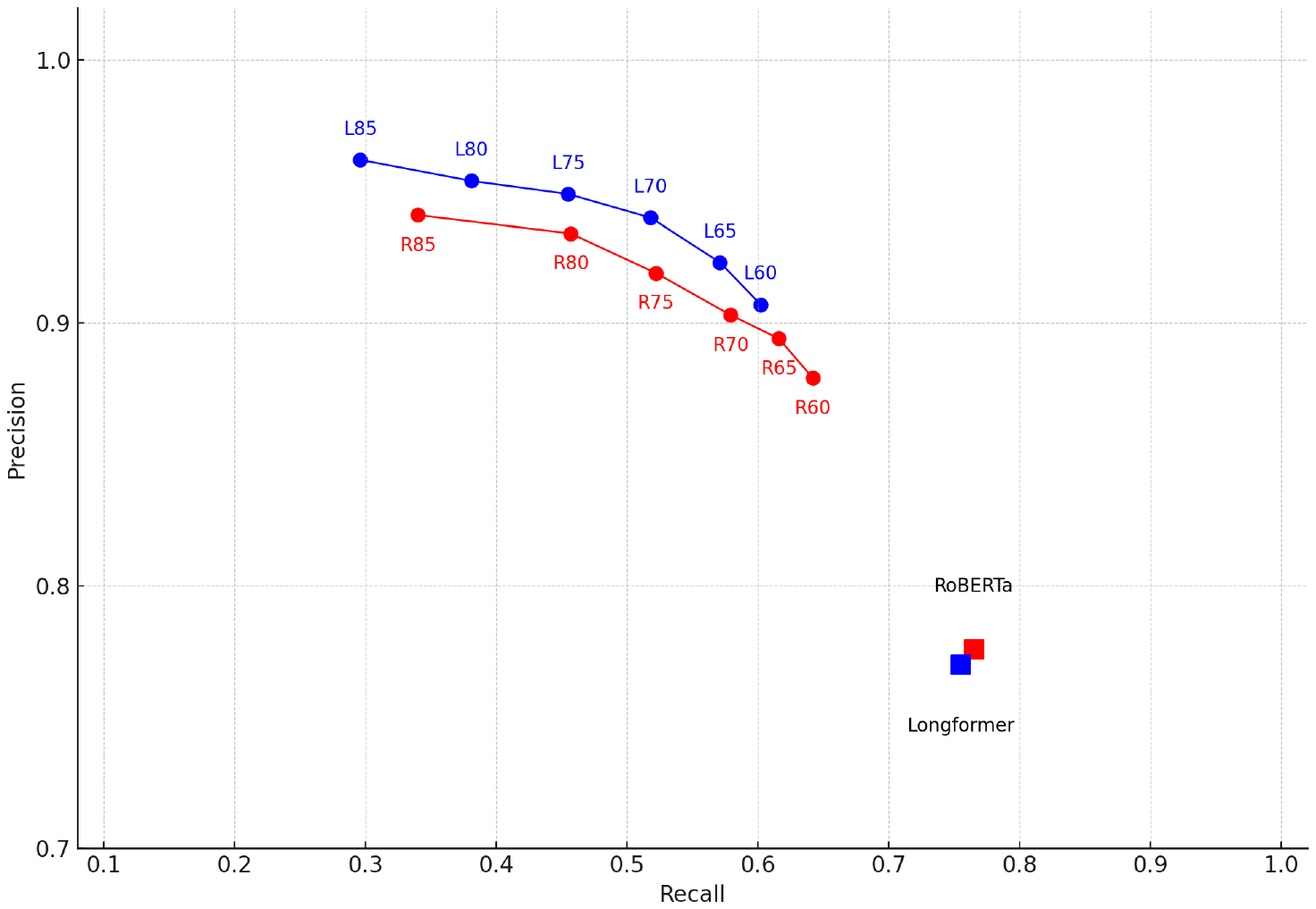}
  \vspace{-0.5em} 
  \caption{Precision–recall trade-off curves for Longformer and RoBERTa using confidence thresholds.}
  \label{fig:LR_recallPrecision}
  \vspace{-0.8em} 
\end{figure}
Ultimately, our goal is develop classification models that can be used to rapidly populate knowledge resources with task-specific information about real-world entities. For SIC code classification task, our best classification model achieved 82.7\% precision, which is  good. But ideally we would like even higher precision so that some instances can be labeled fully automatically without the need for human inspection. 

Toward this end, we experimented with  confidence thresholds to effect a trade-off between precision and recall. This approach considers the confidence of each label predicted by the classifier and only assigns the label if the confidence  exceeds a threshold. Our best classifiers used  GPT-4o mini,  but it  does not provide built-in support to extract confidence scores or log probabilities.
So we ran these experiments with  RoBERTa and Longformer.

Figure~\ref{fig:LR_recallPrecision} displays the results. The squares in the lower right corner show the performance of RoBERTa and Longformer when labeling every instance (no threshold). The other data points show their performance using threshold values ranging from .60 to .85 in increments of five. 
Longformer's confidence scores proved more reliable than RoBERTa's, producing stronger high-precision results.  

This strategy successfully produced high precision ($>$ 90\%) classification while sacrificing some recall. For example, with a threshold of 0.85, Longformer achieved 96\% precision with around 30\% recall; using 0.60, it reached 91\% precision with 60\% recall. These high-precision classifiers may support fully automatic knowledge population for the most confident predictions.

\vspace{-0.7em}
\section{Conclusion}
We introduced a framework that, given only entity names and their corresponding gold labels as input, can automatically generate descriptive text for those entities, which can then be used to train a classifier. The gold labels are provided only for model training and are not used during text acquisition, allowing the framework to operate independently of any pre-existing structured text-based resources. Our approach leverages web search and large language models to automatically acquire task-relevant text. We evaluated our framework on two classification tasks from distinct domains. The healthcare taxonomy code classification task is entirely different from the SIC code classification task, demonstrating the robustness of our proposed framework. Overall, our framework offers a scalable solution for developing entity classification models across diverse real-world tasks, including those involving lesser-known entities.



\section{Ethics Statement}
All healthcare providers included in our benchmark are based in the United States. We obtained the provider names and their corresponding taxonomy codes from the National Plan and Provider Enumeration System (NPPES), maintained by the Centers for Medicare \& Medicaid Services (CMS). The NPPES registry is publicly accessible and downloadable in the U.S., and the data are released by CMS as publicly available records. Healthcare providers submit this information themselves as part of the National Provider Identifier (NPI) registration process, which is required to become HIPAA-covered healthcare providers in the United States.

Our healthcare provider dataset includes Google search snippets retrieved using SerpAPI and large language model (LLM)-generated text. To avoid redistributing third-party content, we do not release the raw Google snippets or the LLM-generated text. Instead, we provide detailed instructions in our GitHub repository describing how the dataset can be reconstructed using the same retrieval procedure, model version, and prompting setup. This approach ensures transparency and reproducibility without redistributing third-party or model-generated content.

All experiments were conducted for research purposes only. The dataset does not contain sensitive attributes beyond publicly available professional information. We believe these measures ensure compliance with scientific integrity standards while minimizing potential ethical risks related to data redistribution.


\section{Acknowledgements}
This research was supported in part by the ICICLE project through NSF award OAC-2112606. We thank Tianyu Jiang for helpful clarifications on their publicly released codebase, which facilitated our reproduction of the results reported in their work.



\section{Bibliographical References}\label{sec:reference}
\bibliographystyle{lrec2026-natbib}
\bibliography{custom}

\section{Language Resource References}
\label{lr:ref}
\bibliographystylelanguageresource{lrec2026-natbib}
\bibliographylanguageresource{languageresource}

\appendix

\section{Appendix}
\label{sec:appendix}
\begin{figure}[H]
  \centering
  \includegraphics[width=\columnwidth]{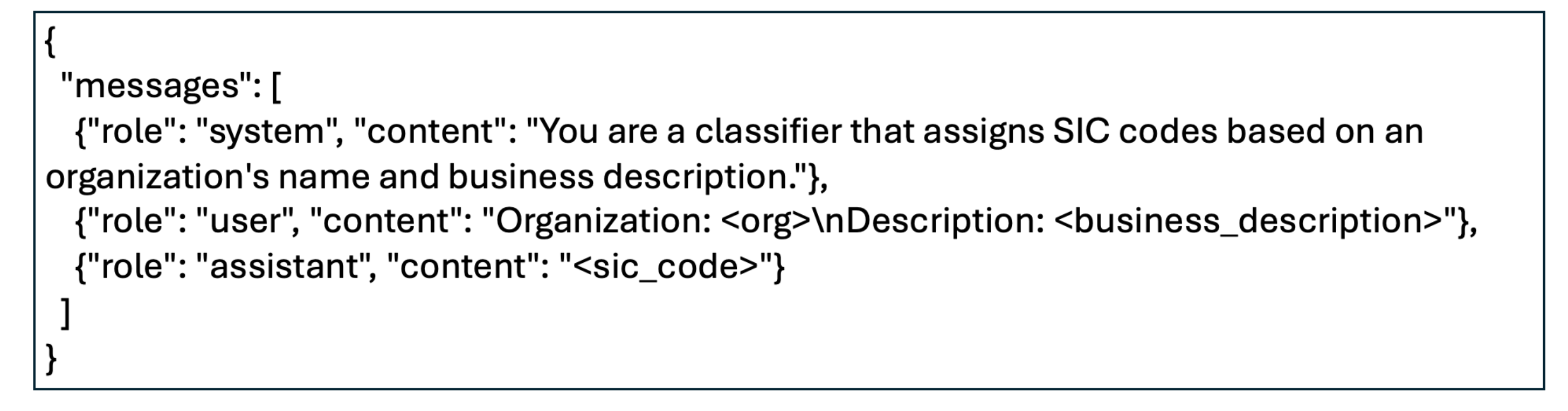}
\caption{Example of the train and dev instance format used to fine-tune \GPTFourMini{}. Each instance consists of a system instruction, a user message combining the organization name and its business description, and an assistant response containing the target SIC code.}
  \label{fig:finetuneGPT-trainDev}
\end{figure}
\begin{figure}[h]
  \centering
  \includegraphics[width=\columnwidth]{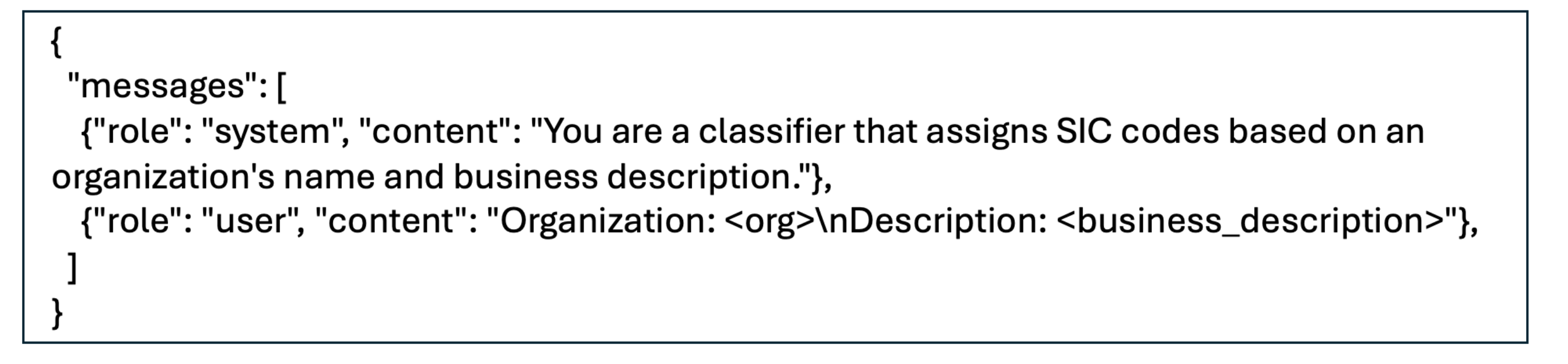}
  \caption{Input format used during inference with the fine-tuned \textsc{GPT-4o mini} model. The model receives a system instruction and a user message containing the organization name and its business description. No gold label is provided during inference.}
  \label{fig:finetuneGPT-inference}
\end{figure}

\end{document}